\pdfoutput=1

\documentclass[11pt]{article}

\usepackage[preprint]{acl}

\usepackage{times}
\usepackage{latexsym}

\usepackage[T1]{fontenc}

\usepackage[utf8]{inputenc}

\usepackage{microtype}

\usepackage{inconsolata}

\usepackage{graphicx}
\usepackage{amssymb}
\usepackage{multirow}

\definecolor{ccd}{rgb}{0.843, 0.909, 0.945}
\definecolor{ours}{rgb}{0.898, 0.886, 0.941}
\definecolor{psychat}{rgb}{0.937, 0.937, 0.8}
\definecolor{camel}{rgb}{0.89, 0.933, 0.905}
\definecolor{nofinetune}{rgb}{0.972, 0.917, 0.917}

\usepackage{caption} 
\title{DiaCBT: A Long-Periodic Dialogue Corpus Guided by Cognitive Conceptualization Diagram for CBT-based Psychological Counseling}

\author{Yougen Zhou\textsuperscript{1$^\ast$}, Ningning Zhou\textsuperscript{3\thanks{These authors contributed equally to this work.}}, Qin Chen\textsuperscript{2\thanks{Corresponding Author}}, Jie Zhou\textsuperscript{2}, Aimin Zhou\textsuperscript{1}, Liang He\textsuperscript{2} \\
\textsuperscript{1}{Shanghai Institute of Artificial Intelligence for Education, East China Normal University}
\\
\textsuperscript{2}School of Computer Science and Technology, East China Normal University
\\
\textsuperscript{3}School of Psychology and Cognitive Science, East China Normal University
}

\begin{document}
\maketitle
\begin{abstract}
Psychotherapy reaches only a small fraction of individuals suffering from mental disorders due to social stigma and the limited availability of therapists. Large language models (LLMs), when equipped with professional psychotherapeutic skills, offer a promising solution to expand access to mental health services. However, the lack of psychological conversation datasets presents significant challenges in developing effective psychotherapy-guided conversational agents. In this paper, we construct a long-periodic dialogue corpus for counseling based on cognitive behavioral therapy (CBT). Our curated dataset includes multiple sessions for each counseling and incorporates cognitive conceptualization diagrams (CCDs) to guide client simulation across diverse scenarios. To evaluate the utility of our dataset, we train an in-depth counseling model and present a comprehensive evaluation framework to benchmark it against established psychological criteria for CBT-based counseling. Results demonstrate that DiaCBT effectively enhances LLMs’ ability to emulate psychologists with CBT expertise, underscoring its potential for training more professional counseling agents.
\end{abstract}

\section{Introduction}

\begin{figure}[t]
  \includegraphics[width=\columnwidth]{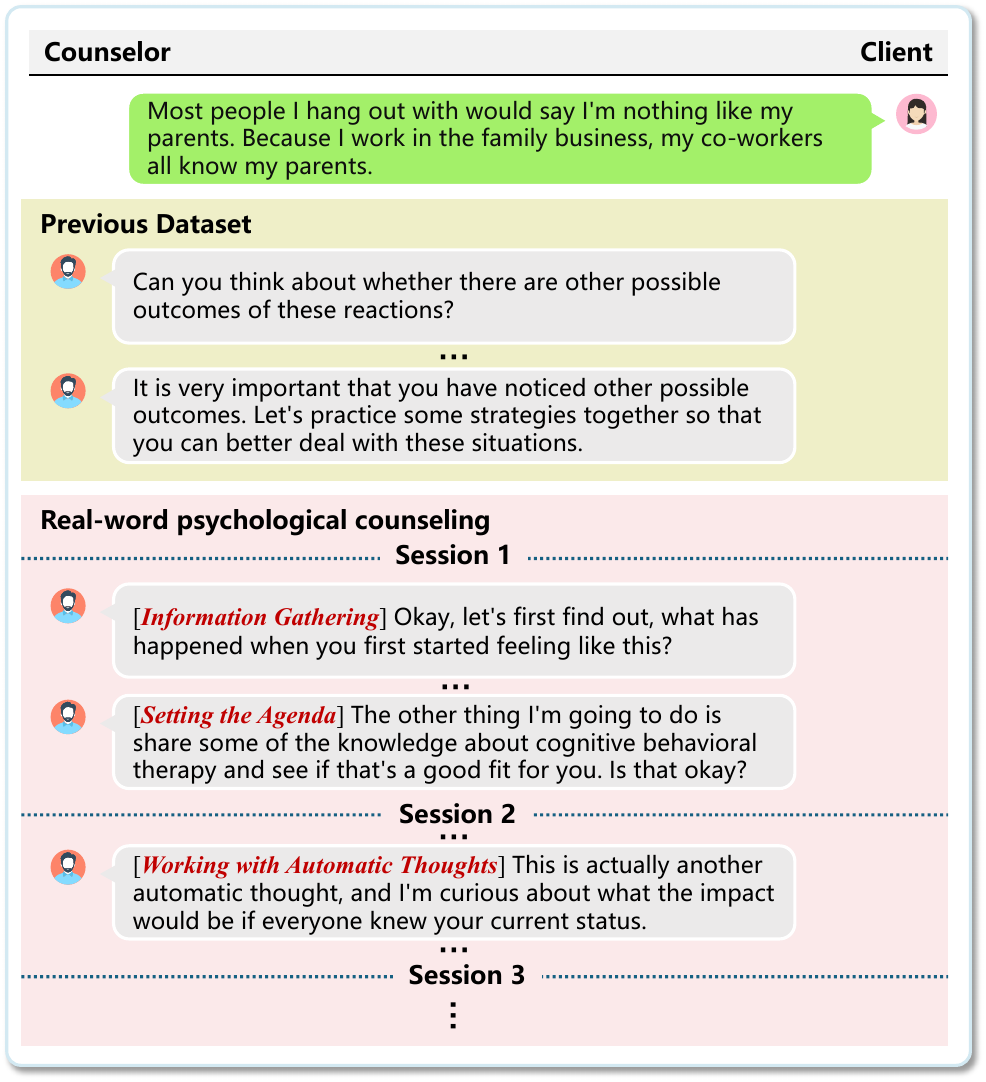}
  \caption{A dialogue comparing a previous counseling dataset to a real-world scenario. Previous datasets tend to solve all psychological problems in a single counseling session using one CBT technique, whereas real-world counseling involves multiple sessions, depending on the nuanced progression of therapeutic strategies.}
  \label{fig:example}
\end{figure}

Mental health disorders are a significant and widespread public health concern, affecting an estimated one in eight people globally (World Health Organization, 2022\footnote{\url{https://www.who.int/campaigns/world-mental-health-day/2022}}). Despite this, access to mental health services remains limited, largely due to social stigma and a shortage of therapists \cite{freeman2022world, white2001receiving}. To address these barriers, there is growing interest in using automated conversational agents as alternative tools for mental health support \cite{ali2020virtual, sabour2023chatbot, sharma2023human}. Whereas, generating human-like responses as the psychologist in automated systems remains a major challenge.

Recently, the integration of large language models (LLMs) into counseling systems presents a promising way \cite{stade2024large, bubeck2023sparks}. While LLMs demonstrate remarkable capabilities in generating human-like responses, they tend to provide generic advice or information \cite{chiu2024computational}. Existing solutions attempt to model the psychological theory as rules or tree flows to prompt LLMs to mimic psychological counselors for professional response generation \cite{mousavi2021would, das2022conversational}. Whereas, the expertise still remains constrained due to the insufficient training of the backbone models in psychological counseling \cite{raile2024usefulness}. Moreover, psychological counseling heavily relies on case practice, while the counseling process data is severely limited by the sensitive nature of therapist-client interactions \cite{harrigian2021state, perez2018analyzing}. 
 

An alternative solution is to construct synthetic dialogues that can resemble real-world counseling processes in diverse scenarios. Previous efforts primarily focus on generating dialogues between therapists and clients based on certain mental problems \cite{qiu-etal-2024-smile, na2024cbt}. Whereas, there is a large gap between the synthetic data and the real-world counseling process. As shown in Figure \ref{fig:example}, the existing synthetic data assumes that all psychological problems can be solved in one counseling session, thus one dialogue mimics the whole counseling process for one case, which is coarse-grained and can not well reflect the difference of counseling strategy use in each session. In practice, real-world psychological counseling often involves long-periodic dialogues with multiple sessions, depending on the difficulty and nuanced progression of therapeutic strategies \cite{curwen2018brief}. Moreover, most client simulation relies on background descriptions such as occupations, experiences, and problems \cite{lee2024cactus}, which can not well model the mental portrait and limits the utterance generation with specific mental problems for each client \cite{wang-etal-2024-patient}.

To address the above challenges, we present \textbf{DiaCBT}, a long-periodic dialogue corpus guided by cognitive conceptualization diagram (CCDs) for psychotherapy counseling based on cognitive behavioral therapy (CBT) \cite{beck2020cognitive}. DiaCBT simulates the entire process of CBT, consisting of multiple sessions that feature interactive transcripts annotated with CBT strategies. To better simulate clients, we construct structured CCDs across diverse scenarios to guide the utterance generation of clients with specific mental problems. We conduct experiments by fine-tuning models for therapy-based response generation using DiaCBT, and present a comprehensive evaluation framework with established psychological criteria for CBT-based counseling. Both automatic and human evaluations demonstrate that DiaCBT significantly enhances LLMs’ ability to emulate psychologists with CBT expertise, highlighting its potential for training professional counseling agents. The main contributions of our work are as follows:
\begin{itemize}
    \item We create a long-periodic CBT-based counseling dialogue dataset, which well mimics the whole process of cognitive behavioral therapy in practice and models clients with structured cognitive conceptualization diagrams during multiple sessions.
    \item We train a psychological counseling model based on DiaCBT, which enhances CBT-specific skills by integrating strategies for in-depth questioning to effectively reframe clients' cognitive distortions.
    \item We present a comprehensive evaluation for psychological counseling and conduct extensive experiments, demonstrating the great advantages of our dataset and counseling model over the baselines.
\end{itemize}

\section{Related Work}

In this section, we review studies about conversational psychotherapy AI, covering both therapy-based conversational systems and datasets.

\subsection{Therapy-based Conversational Systems}
Conversational systems for psychotherapy aim to assess an individual’s mental state and enhance self-awareness through effective communication techniques. These systems often rely on empathetic conversations to alleviate emotional distress, crafting responses that are relevant to the client’s statements \cite{ma2020survey, zhou2018emotional, lubis2018eliciting, raamkumar2022empathetic, li2022knowledge, gao2021improving, shen2021constructing, liu2021towards, cheng2022improving}. The emergence of large language models (LLMs) is revolutionizing therapy-based conversational systems \cite{liu2023chatcounselor, chen2023soulchat, jo2023understanding, wei2024leveraging}. Researchers are increasingly integrating professional counseling strategies into these models to enhance their effectiveness \cite{hsu2023helping, chen2023llm, na2024cbt}, and primarily center around emotion analysis and symptom identification. However, conversational agents from a cognitive-behavioral perspective remain underexplored. Therefore, our goal is to improve the cognitive behavioral therapy skills of conversational agents  in therapeutic settings.

\begin{figure*}[t]
  \includegraphics[width=\linewidth]{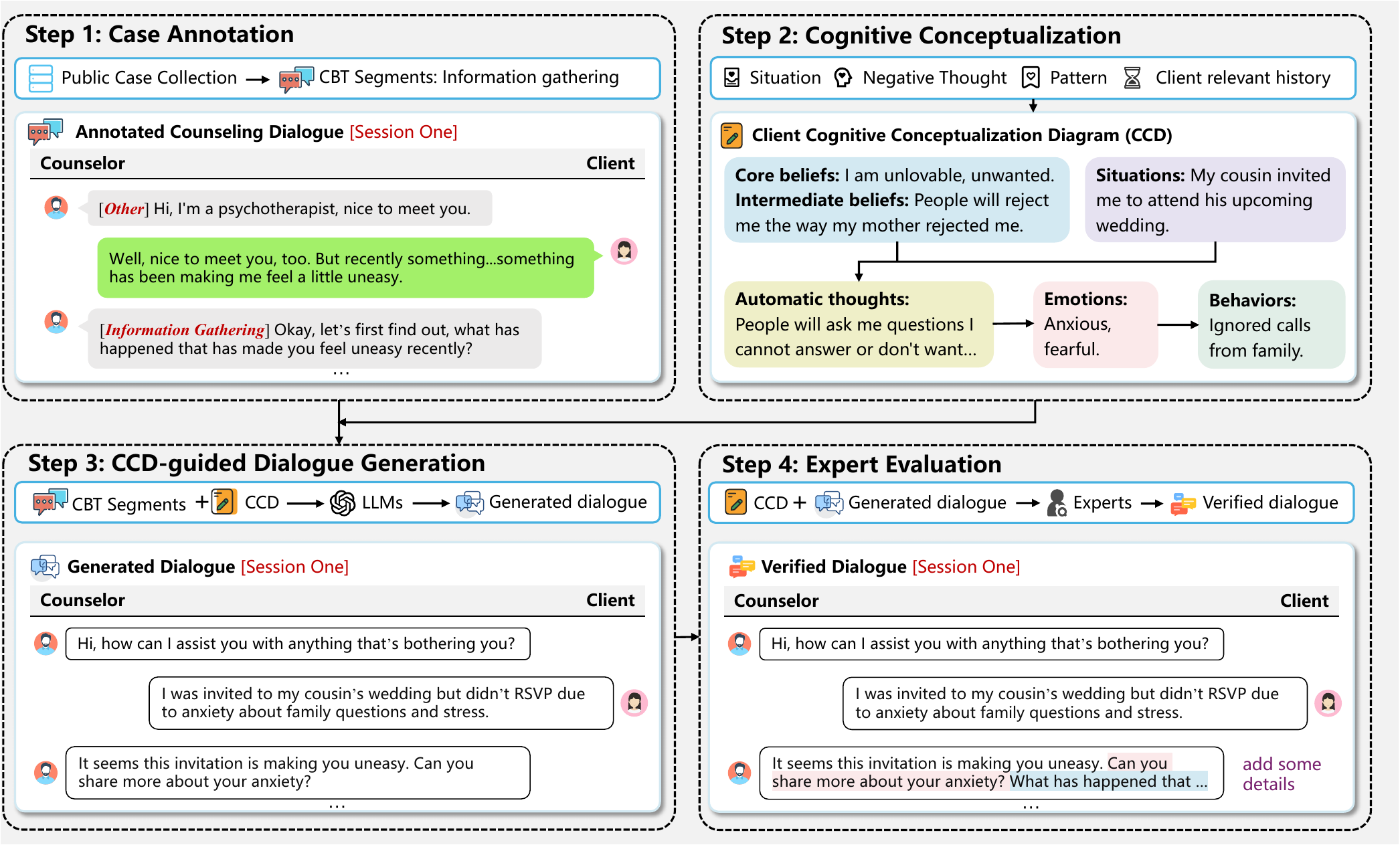}
  \caption{Data creation pipeline of DiaCBT. Step 1: Human annotators collect public cases, annotate CBT sessions, and identify therapist strategies. Step 2: Then, LLMs generate cognitive conceptualization diagrams (CCDs) to enrich client profiles. Step 3: Full client-therapist chat sessions are built using CCDs, CBT sessions with a scripted framework. Step 4: Generated cases are reviewed and edited by experts.}
  \label{fig:main}
  \vspace{-1.5em}
\end{figure*}

\subsection{Psychotherapy Dialogue Datasets}
\citet{perez2018analyzing} introduce a dataset of high- and low-quality counseling conversations that include counseling skills, collected from publicly available sources. Several studies have also leveraged social media platforms for mental health research \cite{sharma2020computational}. For instance, \citet{rashkin2019towards} and \citet{mousavi2021would} develop corpora of empathetic and therapeutic dialogues from real-life interactions, while \citet{liu2021towards} create an emotional support conversation dataset with annotations based on Helping Skills Theory. Additionally, \citet{yao2022d4} design a 3-phase procedure simulating doctor-patient dialogues for depression diagnosis. \citet{wu2022anno} introduced AnnoMI, the first publicly accessible dataset of professionally transcribed and expert-annotated therapy dialogues. Related works have also explored using LLMs to synthesize specific dialogue data, offering a promising alternative to real data \cite{na2024cbt, zhang-etal-2024-cpsycoun, lee2024cactus, xiao-etal-2024-healme}. However, these studies predominantly focus on emotional disorders or depression, limiting their ability to support the broader range of strategies. Additionally, these conversations tend to be short, lacking the depth and multi-stage complexity of face-to-face clinical psychotherapy.

\section{Dataset Construction}

Previous studies have sourced psychological counseling dialogues from online welfare counseling platforms or generated data through crowdsourcing simulations \cite{li-etal-2023-understanding, yao2022d4}. However, due to the sensitive nature of clinical records, access to real clinical data is restricted. To address this, we design a specialized task to generate dialogues. Unlike prior approaches that rely on large-scale recruitment of online volunteers, our annotation process combines the efforts of large language models with contributions from trained volunteers and domain experts. As illustrated in Figure \ref{fig:main}, our methodology comprises four key steps: case annotation, cognitive conceptualization, CCD-guided dialogue generation, and expert evaluation, each of which will be further detailed.

\subsection{Case Annotation}

\paragraph{Public case source.} Our case collection consists of transcripts of cognitive behavioral therapy (CBT) sessions sourced from the American Psychological Association (APA) website\footnote{\url{https://www.apa.org/pubs/databases/psyctherapy/}} and related books\footnote{A CBT textbook named \emph{Dispelling the Fog of Belief} selected by our domain expert. Our expert is a licensed CBT therapist with 300+ hours of counseling experience.}. We selected these cases for two key reasons: (1) their strong relevance to the clinical setting and (2) their widespread use in training novice CBT counselors. A total of 53 transcripts were collected, which can be divided into 33 sessions in 6 complete consultation processes. In collected sessions, the avg. utterance numbers and avg. tokens are 279.5 and 6759.2, respectively. Nearly the entire samples are comprised of more than 100 utterances, which significantly surpasses the input limit of many language backbones and considerably more turns than typical for previous counseling dialogue datasets.
\begin{table}[]
\centering
\resizebox{\columnwidth}{!}{
\begin{tabular}{lccc}
\hline
\textbf{Category} & \textbf{Total} & \textbf{Client} & \textbf{Therapist} \\
\hline
Cases & 108 & - & - \\
Avg. utterances per case & 264.87 & 133.92 & 130.92 \\
Avg. tokens per case & 6253.20 & 3222.20 & 3031.00 \\
Dialogue Sessions & 540 & - & - \\
Avg. utterances per session & 52.97 & 26.78 & 26.18 \\
Avg. tokens per utterance & 23.80 & 24.12 & 23.48 \\
\hline
\end{tabular}
}
\caption{Statistics of DiaCBT.}
\label{tab:case_statistics}
\end{table}
\paragraph{Data annotation target.} CBT, a leading psychological treatment, improves mental health through evidence-based communication. In our work, we incorporate 14 counseling strategies, the detailed definitions of the strategy are in Appendix \ref{sec:label_system}. For annotation, annotators\footnote{Our annotators are graduate students in psychology with foundation knowledge in CBT. They have undergone a comprehensive training program on CBT principles and annotation guidelines, and have all passed the annotation competency test.} are provided with conversation transcripts and tasked with identifying the CBT strategies employed by the therapist according to the guidelines provided. While strategies may overlap, we simplify the annotation process by treating it as a single-label task. Annotators focus on annotating the strategy used in the current segment and, in cases of overlap, are instructed to select the primary strategy label.

\subsection{Cognitive Conceptualization}
In addition to the public case library, many psychology-trained individuals have contributed to public datasets through crowdsourcing, particularly for analyzing cognitive distortions and cognitive reframing \cite{wang-etal-2023-c2d2, maddela-etal-2023-training}. While these datasets are not exclusively conversational, their content closely mirrors the thoughts of real clients, making them valuable for enriching client profiles. To this end, we compile various CBT-related datasets and construct cognitive models with LLMs based on information provided in these datasets to accurately simulate backgrounds that resemble those of real patients.

We utilize the Cognitive Conceptualization Diagram (CCD), recognized as a commonly used representation of a patient’s cognitive model in CBT. The CCD includes components for understanding how an individual’s thoughts and beliefs are interconnected and influence emotions and behaviors. We select six key components for formulating a patient’s cognitive model\footnote{Details regarding the rationale for selecting these components are provided in Appendix \ref{sec:ccd}.}. \emph{Core Beliefs} are deeply ingrained perceptions about oneself, others, and the world. \emph{Intermediate beliefs} are the underlying rules, attitudes, and assumptions derived from core beliefs and shape an individual’s thought patterns. An external event or context (a \emph{situation}) may trigger quick, evaluative thoughts without deliberation (\emph{automatic thoughts}) stemming from the beliefs, leading to responses in terms of \emph{emotions} and \emph{behaviors}. In this work, we integrate situation, negative thought, cognitive pattern, and relevant history from the C2D2 \cite{wang-etal-2023-c2d2} and PatternReframe \cite{maddela-etal-2023-training} as contexts for generating CCD-based cognitive models.

\begin{table}
\centering
\resizebox{\columnwidth}{!}{
\begin{tabular}{lcc}
\hline
\textbf{Strategy} & \textbf{Segments} & \textbf{Utterances} \\
\hline
Information Gathering & 303 & 7565\\
Setting the Agenda & 129 & 2383\\
Weekly Review & 249 & 5342\\
Defining Therapeutic Objectives & 78 & 1599\\
Psychoeducation & 168 & 3411\\
Working with Automatic Thoughts & 438 & 10962\\
Motivational Enhancement & 171 & 3445\\
Working with Intermediate and Core Beliefs & 192 & 5398\\
Behavioral Techniques & 195 & 4136\\
Relapse Prevention & 18 & 420\\
Homework Assignments & 168 & 3096\\
Requesting Feedback & 135 & 2721\\
Summarization & 72 & 1484\\
Other & 297 & 5714\\
\hline
\end{tabular}
}
\caption{Data statistics of strategy categorization. We count the number of strategy-coherent segments of each strategy, and the number of utterances.}
\label{tab:data_statistic}
\end{table}

\subsection{CCD-guided Dialogue Generation}
LLMs trained on extensive text corpora can effectively support data collection \cite{zhang-etal-2024-cpsycoun, lee2024cactus, xiao-etal-2024-healme}, which demonstrate knowledge of psychological therapy concepts, including CBT techniques. To this end, we enhance dialogue generation by using a cognitive model about clients and CBT sessions with a scripted framework. We use annotated segments from different CBT sessions as few-shot prompts to guide LLMs in generating conversational flow. Additionally, we prompt LLMs to role-play as clients based on a CCD and follow instructional prompts to reflect the underlying cognitive processes. This approach ensures that therapists in the generated dialogues adhere to the full CBT process, helping clients identify solutions independently rather than simply reframing their thoughts. The instructional template is provided in the Appendix \ref{sec:dialogue_generation}.

\begin{figure}[t]
    \centering
    \includegraphics[width=\linewidth]{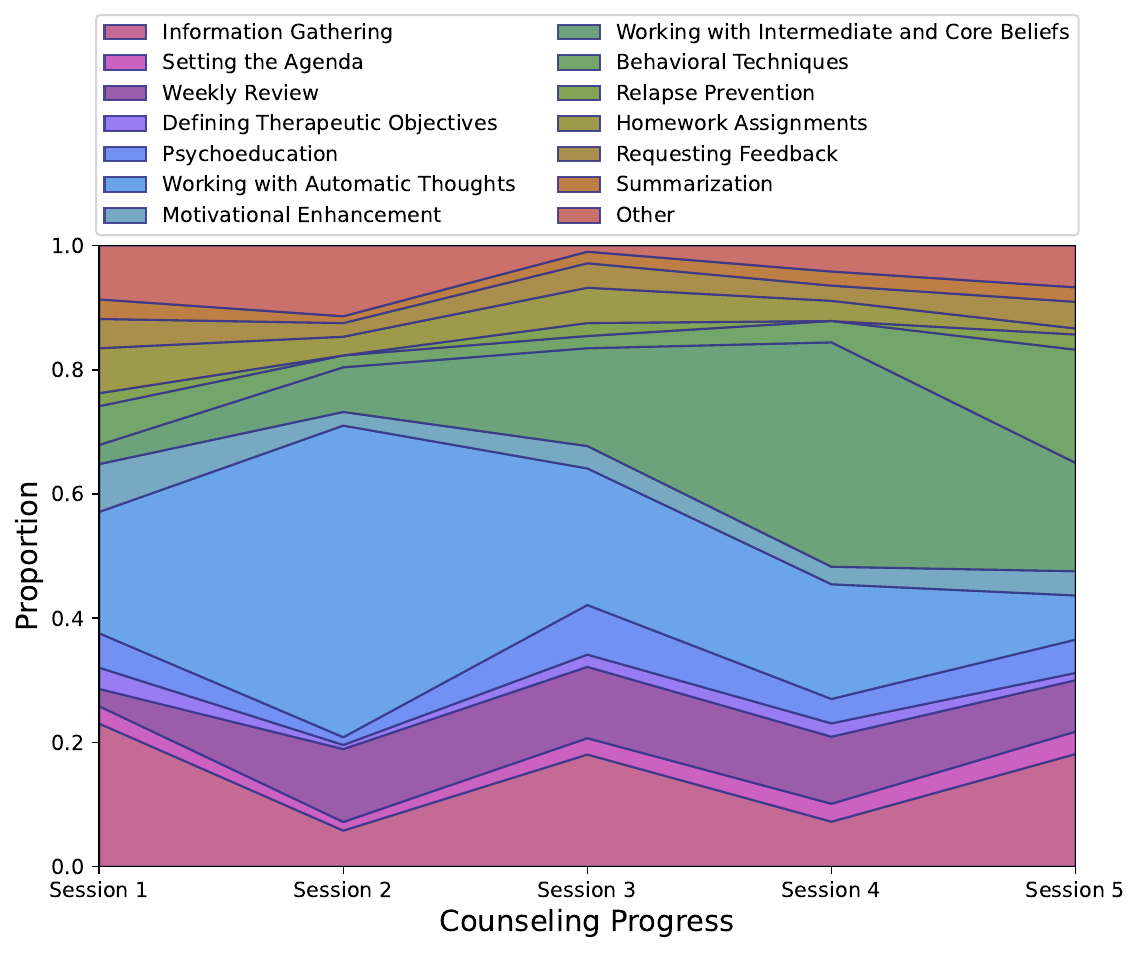}
    \caption{The distribution of strategies used in conversations at each session.}
    \label{fig:strategy_transfer}
\end{figure}
\subsection{Expert Evaluation}
To ensure the data quality, all sessions are then manually screened and we manually annotate a small portion and develop rules, including \textbf{Correctness}, \textbf{Reasonableness}, and \textbf{Situation Diversity}, to distinguish high-quality data from suboptimal data. Further details are provided in the Appendix \ref{sec:label_evaluation}. We also rephrase statements into more natural, colloquial language, especially for question asked by therapist, which LLMs often express too simple. These criteria ensure that the generated dialogues are not only theoretically accurate but also practical and diverse enough to support varied applications in counseling contexts. The final reserved data contains 2613 segments from the raw 3600 segments, resulting in an overall retention rate of around 72.58\%. This statistic proves that the psychotherapy ability exhibited by most powerful LLMs is still not satisfactory enough even with delicate prompting, thus further emphasizing the necessity of our construction of DiaCBT.

\section{Data Analysis and Model Training}
\subsection{Data Statistics}
The dataset statistics are displayed in Table \ref{tab:case_statistics}. Our dataset comprises 108 cases, with an average of 264.87 utterances per case and an average of 6253.2 tokens per case. In CBT counseling, each case consists of five sessions, with specific strategy segments used in each session. The data is further segmented into 540 sessions, averaging 52.97 utterances per session. Our dialogues have consistent statistics with those in the original collected examples, which shows that our data is well aligned with real-world scenarios. 

We also present the statistics of other annotations in Table \ref{tab:data_statistic}, showcasing the segment number of each strategy and their utterance number. Our dataset comprises a total of 2613 segments. We observe that some strategies are frequent such as `\emph{Working with Automatic Thoughts}', `\emph{Information Gathering}', and `\emph{Working with Intermediate and Core Beliefs}'. For efficient communication, therapists organize responses based on client content and CBT programs. As a result, strategies present different importance in the form of frequency and length. Automatic thoughts, intermediate beliefs, and core beliefs as components in the client's CCD are also where the therapist needs to work, which is consistent with the principles of CBT.

\subsection{Strategy Distribution}
We also compute the distribution of strategies at different sessions of psychological counseling, as shown in Figure \ref{fig:strategy_transfer}. In CBT, complete psychological counseling is divided into five sessions. However, there is also flexibility in adopting strategies at each session. For instance, in the early session, therapists often use exploratory strategies like `\emph{Information Gathering}' and work on the client's automatic thoughts using `\emph{Working with Automatic Thoughts}'. After understanding the clients' situations, therapists tend to work on the client's intermediate and core beliefs in later sessions, with psychotherapeutic strategies like `\emph{Working with Intermediate and Core Beliefs}' being used more frequently. This is consistent with the principles of CBT \cite{beck2020cognitive}.

\begin{table*}
    \centering
    \resizebox{0.8\linewidth}{!}{
    \begin{tabular}{c|ccccc}
    \hline
    \textbf{Datasets} & \textbf{CBT-based} & \textbf{Multi-turn} & \textbf{Full-session} & \textbf{Strategy} & \textbf{CCD-guided} \\
    \hline
    AnnoMI \cite{wu2022anno} & - & \checkmark & \checkmark & \checkmark & - \\
    SMILECHAT \cite{qiu-etal-2024-smile} & - & \checkmark & - & - & - \\
    CBT-LLM \cite{na2024cbt} & \checkmark & - & - & - & - \\
    CPsyCoun \cite{zhang-etal-2024-cpsycoun} & \checkmark & \checkmark & - & - & - \\
    Healme \cite{xiao-etal-2024-healme} & \checkmark & \checkmark & - & \checkmark & - \\
    Cactus \cite{lee2024cactus} & \checkmark & \checkmark & - & \checkmark & -\\
    \hline
    \textbf{DiaCBT(Ours)} & \textbf{\checkmark} & \textbf{\checkmark} & \textbf{\checkmark} & \textbf{\checkmark} & \checkmark \\
    \hline
    \end{tabular}
    }
    \caption{Comparison of the key features of our work with related datasets.}
    \label{tab:related_datasets}
    \vspace{-1.5em}
\end{table*}

\subsection{Comparison with Related Datasets}
As shown in Table \ref{tab:related_datasets}, our DiaCBT stands out due to its focus on CBT, supporting multi-turn dialogues and full counseling sessions. First, it is the only dataset in the comparison that is CBT-based, includes strategy annotations, and offers a complete counseling process with detailed dialogue turns. Second, DiaCBT incorporates a wide range of psychological strategies and is based on the Cognitive Conceptualization Diagram (CCD). These features make DiaCBT a comprehensive and unique resource for studying CBT in therapeutic contexts.

Previous CBT dialogue datasets have primarily focused on single-turn strategies, with recent efforts extending to multi-turn dialogues. However, these often involve only 2-3 turns or adapt counseling reports without incorporating specific strategies. In contrast, our approach integrates CCD with annotated CBT segments to generate high-quality dialogues, while maximizing the availability of open datasets for researchers. Additionally, DiaCBT includes a wider range of CBT strategies across various stages, going beyond Cactus's focus on automatic thought work to cover more techniques.

\subsection{Model Training}
We fine-tune an LLM on DiaCBT, enabling it to: (1) select appropriate CBT strategies based on conversation history, (2) generate responses tailored to the client’s input, and (3) generate strategies and responses with given conversation history. The fine-tuning process employs a structured template to serialize $T$-turn dialogues, concatenating task instructions with corresponding conversational context. The training objective is formulated as the minimization of the negative log-likelihood for predicting the therapist's response $u_t^d$ and strategy $s_t$:
\setlength{\abovedisplayskip}{5pt plus 2pt minus 4pt}
\setlength{\belowdisplayskip}{5pt plus 2pt minus 4pt}
\begin{equation}
\min \limits_{\theta} \sum_{t=1}^{T} - \log p_{\theta} \left( s_t, u_t^d \big| I_{task}, h_{t-1}, u_t^c \right)
\end{equation}
where $I_{task}$ denotes the task instruction template serving as a static conditioning signal. $h_{t-1}$ represents the dialogue history, formally defined as: $h_{t-1} = \big( u_0^d, u_0^c, \ldots, u_{t-1}^d, u_{t-1}^c \big)$, where $u_t^d$ and $u_t^c$ indicate the utterances corresponding to the therapist and client respectively in the $t$'th turn. $\theta$ represents the parameters for the LLM. The objective function explicitly conditions therapeutic strategy and response generation on both task-specific instructions and evolving dialogue context.

\begin{table*}
\centering
\resizebox{0.7\linewidth}{!}{
\begin{tabular}{llcccc}
\hline
\textbf{Method} & \textbf{Backbone} & \textbf{AT} $\uparrow$ & \textbf{SR} $\uparrow$ & \textbf{Positive} $\uparrow$ & \textbf{Negative} $\downarrow$ \\
\hline
CpsyCounX & InterLM2-Chat-7B & 6.01 & 34.28\% & 1.179 & -0.910\\
SoulChat & ChatGLM-6B & 6.25 & 38.57\% & 1.135 & -0.992\\
PsyChat & ChatGLM-6B & 8.92 & 72.85\% & 1.445 & \textbf{-1.210}\\
MeChat & ChatGLM-6B & 5.62 & 47.14\% & 1.031 & -0.954\\
CAMEL & LLAMA3-8B-Instruct & 9.42 & 67.14\% & 1.305 & -1.044\\
\hline
Standard & Qwen2.5-7B-Instruct & 6.72 & 44.28\% & 1.112 & -1.122\\
Ours & Qwen2.5-7B-Instruct & \textbf{12.05} & \textbf{77.14\%} & \textbf{1.675} & -1.021\\
\hline
\end{tabular}
}
\caption{The results of metrics and emotion changes for baselines. The best score of each metric is \textbf{in-bold}.}
\label{tab:main_result}
\vspace{-0.5em}
\end{table*}

\setlength{\abovedisplayskip}{5pt plus 2pt minus 4pt}
\setlength{\belowdisplayskip}{5pt plus 2pt minus 4pt}
\begin{table*}
\centering
\resizebox{\linewidth}{!}{
\begin{tabular}{lcccccccccc}
\hline
\multirow{2}{*}{\textbf{Models}} & & \multicolumn{3}{c}{\textbf{General Counseling Skills}} & & \multicolumn{3}{c}{\textbf{CBT-specific Skills}} & & \multirow{2}{*}{\textbf{Sum.}} \\
\cline{3-5} \cline{7-9}
 & & \textbf{Understanding} & \textbf{Interpersonal Eff.} & \textbf{Collaboration} & & \textbf{Guided Discovery} & \textbf{Focus} & \textbf{Strategy} & &\\
\hline
CpsyCounX & & 4.02 & 5.24 & 4.53 & & 4.00 & 4.09 & 4.04 & & 25.92\\
SoulChat & & 4.00 & 4.90 & 4.25 & & 4.00 & 4.00 & 4.00 & & 25.15\\
PsyChat & & \textbf{4.15} & \textbf{5.84} & \textbf{5.63} & & \underline{4.07} & 4.05 & 4.01 & & \underline{27.75}\\
MeChat & & 4.01 & 4.86 & 4.25 & & 4.00 & 4.04 & 4.01 & & 25.17\\
CAMEL & & 4.00 & \underline{5.82} & 4.73 & & 4.02 & \textbf{4.38} & 4.06 & & 27.01\\
\hline
Standard & & 4.02 & 5.70 & 4.35 & & 4.04 & 4.01 & \underline{4.08} & & 26.20\\
Ours & & \underline{4.07} & 5.35 & \underline{5.57} & & \textbf{5.03} & \underline{4.31} & \textbf{4.42} & & \textbf{28.75}\\
\hline
\end{tabular}
}
\caption{The results on general counseling and CBT-specific skills for baselines. The best score of each metric is \textbf{in-bold}, while the second best score is \underline{underlined}.}
\label{tab:ctrs}
\vspace{-1.5em}
\end{table*}

\section{Experiments}

In this work, we propose a computational evaluation framework to evaluate CBT-based psychological counseling agents both in the counselor’s abilities and in the client’s psychological changes.

\subsection{Computational Evaluation Framework}

To evaluate an LLM acting as a therapist, we focus on assessing its responses in supporting clients with mental health challenges. Recruiting real clients with mental health issues and asking them to interact with LLM therapists would pose significant ethical risks. Instead, we leverage client CCDs to simulate clients, facilitating a full, multi-turn conversational session between LLMs therapist and simulated clients.

\paragraph{Client Simulating.} We use cases from the C2D2 \cite{wang-etal-2023-c2d2} to generate new CCDs. C2D2, a Chinese cognitive distortion dataset, contains 7,500 instances categorized into seven cognitive distortion types. We randomly select 20 cases from each distortion label, totaling 140 cases. Each case includes a distortion type, an emotion type, a negative thought, and a situation description, which are used to initialize CCDs. Each CCD is fed into GPT-4o to simulate the client talking to the therapist, maintaining the same conversation style, life events, and emotions. Conversations begin with the therapist and continue until the client outputs an end token. Detailed prompts for client simulating are provided in the Appendix \ref{sec:client_simulate}.

\paragraph{Evaluation Metrics.} Previous studies typically evaluate turn-level performance based on fixed reference responses. Differently, when it comes to the evaluation of counseling dialogue systems, it would be more appropriate to focus on dialogue-level performance. To this end, for automatic evaluation, we employ two key metrics: the \textbf{average turn (AT)} and the \textbf{success rate (SR)} \cite{he-etal-2024-planning, deng2024plug}. AT measures goal completion efficiency by calculating the average number of turns, while SR measures goal completion effectiveness by computing the success rate of achieving the goal within a predefined maximum number of turns. Furthermore, following \citet{lee2024cactus}, we use the Positive and Negative Affect Scale (PANAS) \cite{watson1988development} to assess the effectiveness of counseling from the client’s perspective, measuring changes in positive and negative emotions before and after the session. Finally, the Cognitive Therapy Rating Scale (CTRS) \cite{aarons2012adaptation} is used to evaluate general counseling and CBT-specific skills of LLM therapists. Detailed information is provided in Appendix \ref{sec:mectric}.

\subsection{Experimental Setup}

The DiaCBT dataset is split into training and validation sets at an 8:2 ratio. The training uses LoRA to fine-tune the Hugging Face implementation of Qwen2.5-7B-Instruct \cite{yang2024qwen2} and train it for 3 epochs using the AdamW optimizer with learning rate 1e-4 and batch size 32 in a single NVIDIA A800 GPU. We set the dimension of low-rank matrices to 64 and alpha to 16. 

We compare our model with the recent advanced baselines: 1) CpsyCounX \cite{zhang-etal-2024-cpsycoun} is fine-tuned on a report-based multi-turn dialogue dataset for psychological counseling; 2) SoulChat \cite{chen2023soulchat} is fine-tuned on a multi-turn empathetic dialogue dataset to generate empathetic responses; 3) PsyChat \cite{qiu2024psychat} is a client-centric dialogue system that provides psychological support based on client behavior recognition and counselor strategy selection; 4) MeChat \cite{qiu-etal-2024-smile} is trained on a multi-turn mental health support dialogue dataset extended from real psychological mutual assistance QA; 5) CAMEL \cite{lee2024cactus} is a CBT-based agent, which uses planning with CBT techniques for psychological counseling.

\subsection{Results}

\paragraph{Main Result.} The metrics in Table \ref{tab:main_result} indicate that training with DiaCBT improves both the length of counseling sessions and the success rate of interactions. Compared to the standard Qwen2.5-7B-Instruct, our model demonstrates enhanced counseling skills, highlighting DiaCBT’s effectiveness. PsyChat, trained with counselor strategy selection tasks, outperforms other baselines, underscoring the value of therapist strategy annotations in improving LLM therapist performance. These results further validate DiaCBT as a high-quality dataset that closely mirrors real counseling scenarios.

\paragraph{Evaluation from the Client’s Perspective.} We use PANAS to assess efficacy through changes in client emotions. Table \ref{tab:main_result} shows that DiaCBT effectively enhances positive emotions but is less effective in reducing negative emotions. This may stem from the focus of CBT on improving mental health and emotional regulation through evidence-based communication. Specifically, the strategy `\emph{Working with Automatic Thoughts}' guides clients to explore thought patterns from various perspectives rather than directly altering them.

\paragraph{Evaluation from the Counseling Skills.} The results are presented in Table \ref{tab:ctrs}. Each skill is evaluated using three criteria, with scores ranging from 0 to 6. While training with DiaCBT yields a marginal performance dip in general counseling competencies, it achieves improvements in CBT-specific skills, particularly in Guided Discovery and Strategic Implementation. Training with DiaCBT achieves the highest scores, reflecting the alignment with evidence-based CBT principles, where active client collaboration and structured cognitive-behavioral exploration are prioritized over non-directive counseling approaches.


\subsection{Human Evaluation}
Beyond automatic evaluation, we also explore how humans perceive the generated responses. We conduct a human evaluation on 140 generated dialogues, where three annotators are asked to compare the generated responses from DiaCBT with two competitive baselines: (1) PsyChat and (2) CAMEL. We evaluate the responses based on three main criteria: \textbf{Relevance}, \textbf{CBT Style Measure}, and \textbf{Helpfulness}. The instructions for annotators are provided in the Appendix \ref{sec:human}. As shown in Table \ref{tab:human_evaluate}, DiaCBT outperforms other baselines in nearly all aspects of the human evaluation, as well as the overall evaluation, except for Relevance, where CAMEL achieves a higher win rate. Qualitative case study details for different models are provided in the Appendix \ref{sec:case_study}. We observe that CAMEL excels in providing detailed counseling strategy questions, most of which are closed questions, contributing to its strong relevance. However, for more advanced counseling, the system should go beyond merely asking yes/no questions and actively explore and help resolve the client’s negative thoughts.

\subsection{Further Analysis}
\begin{table}[]
\centering
\resizebox{0.72\linewidth}{!}{
\begin{tabular}{lccccc}
\hline
\multirow{2}{*}{\textbf{Ours vs.}} & \multicolumn{2}{c}{\textbf{CAMEL}} & & \multicolumn{2}{c}{\textbf{PsyChat}} \\
\cline{2-3} \cline{5-6}
 & \textbf{Win} & \textbf{Lose} & & \textbf{Win} & \textbf{Lose} \\
\hline
Relevance & 51 & 89 & & 66 & 73 \\
CBT Style & 76 & 53 & & 92 & 17 \\
Helpfulness & 56 & 24 & & 96 & 28 \\
\hline
\textbf{Overall} & 88 & 34 & & 106 & 24 \\
\hline
\end{tabular}
}
\caption{Human evaluation results. Ties are not shown.}
\label{tab:human_evaluate}
\end{table}

\begin{figure}
    \centering
    \includegraphics[width=\linewidth]{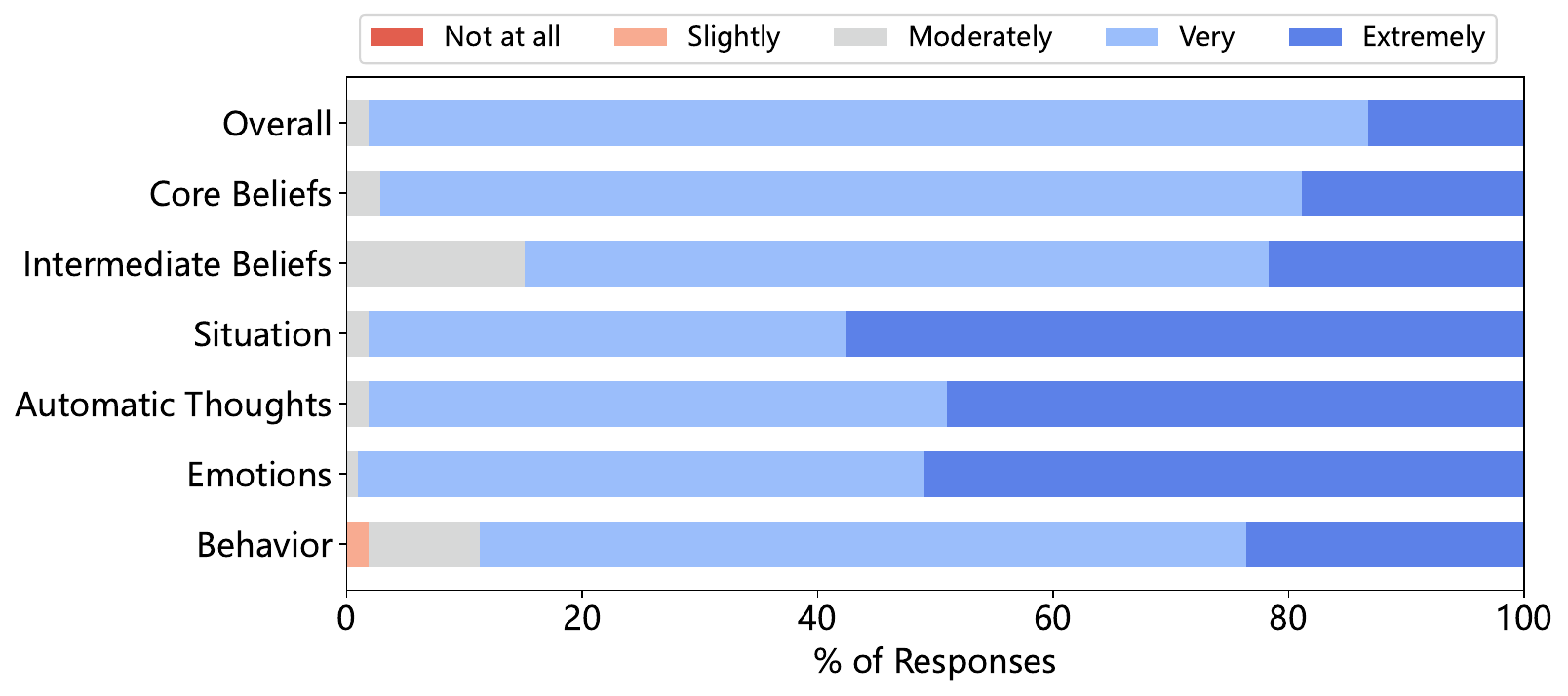}
    \caption{The distribution of ratings. Experts rate over 80\% of the simulated clients as very to extremely accurate for each component of the cognitive model.}
    \label{fig:ccd_accuracy}
\end{figure}

\paragraph{Analysis of LLMs as Client Simulators.} Simulated clients are expected to embody their assigned roles based on specific profiles. Building on prior studies utilizing LLMs as client simulators \cite{wang-etal-2024-patient}, we assess the accuracy of these simulated clients in reflecting their underlying CCDs during multi-turn conversations. Three experts evaluate the simulated clients’ overall accuracy and the accuracy of each component in CCD. Figure \ref{fig:ccd_accuracy} displays the distribution of ratings. The results are promising: overall, simulated clients based on CCDs are rated as very accurate. For each of the six components of CCD, the simulated clients receive average ratings ranging from very to extremely accurate. These findings validate the reliability of leveraging LLMs as client simulators, demonstrating their capability to accurately emulate diverse cognitive processes in counseling scenarios.
\begin{table}[]
\centering
\resizebox{\columnwidth}{!}{
\begin{tabular}{l|ccc|c}
\hline
\textbf{Metrics} & \textbf{Standard} & \textbf{PsyChat} & \textbf{CAMEL} & \textbf{Ours} \\
\hline
Avg. turns & 6.72 & 8.92 & 9.42 & 12.05\\
Avg. doctor utterance length & 61.15 & 53.39 & 43.04 & 31.13 \\
Avg. patient utterance length & 46.74 & 46.97 & 41.89 & 49.44 \\
\hline
Avg. question & 4.46 & 7.18 & 7.66 & 9.04 \\
Avg. in-depth question & 2.75 & 5.85 & 6.68 & 7.41 \\
\hline
\end{tabular}
}
\caption{Automatic evaluation results of LLMs therapist. In-depth questions encourage clients to elaborate on their thoughts, feelings, or experiences in detail.}
\label{tab:question_style}
\end{table}

\paragraph{Analysis of Question Style.} LLM therapists should guide clients to find solutions independently through reflective questioning. To assess what models learned from DiaCBT, we calculated several automatic metrics from 140 dialogue histories, with results presented in Table 7. We observe that the non-finetuned model has the fewest average turns and the highest amount of words per turn. This model often provides general advice, reflecting a lower level of expertise as an LLM therapist. In contrast, PsyChat, CAMEL, and ours demonstrate increased average turns and questions asked, indicating that training with counseling dialogues enhances questioning capabilities. Additionally, our model exhibits a more in-depth questioning style, better aligning with the interactive nature of clinical counseling. However, all model exhibits a slight preference bias in strategy use. For example, the strategy of psychoeducation is less used compared to collected sessions. This poses a significant challenge in mitigating strategy bias when guiding LLMs in long-periodic CBT-based counseling.


\section{Conclusions}

In this study, we construct a CBT-based dialogue dataset (DiaCBT) across various sessions, guided by cognitive conceptualization diagrams and CBT flow. Extensive experiments validate the great potential of our dataset for CBT-based psychological counseling. Specifically, the model fine-tuned on DiaCBT outperforms recent advanced psychotherapy conversational agents in reframing clients' cognitive distortions. It is also notable that our method enhances CBT-specific skills by integrating strategies for in-depth questioning, helping clients find solutions independently. In the future, we plan to explore a broader range of psychotherapy techniques and other counseling scenarios.


\section*{Limitations}

In real counseling sessions, each session typically lasts around 45 minutes, with about five sessions focusing on specific topics. While our dataset involves longer multi-turn interactions compared to others, it remains significantly shorter than actual counseling sessions. Future work should aim to include longer conversations and multi-session interactions to better emulate counseling scenarios.


\section*{Ethical Considerations}

\subsection*{Data Privacy}

To preserve privacy and uphold ethical integrity, we adhered strictly to established data protocols during the case collection phase, ensuring that no cultural bias was introduced while migrating data styles. In generating dialogues, we avoided using real client data for simulated counseling scenarios. Instead, we relied on publicly available datasets curated explicitly for research purposes. These datasets were constructed through crowdsourcing information from psychological experts rather than from actual clients, thereby mitigating ethical concerns regarding personal identification and confidentiality breaches. The information provided by psychological experts was generalized, ensuring that it did not reflect any specific individual’s psychological profile. This approach maintained the ethical standards required for data usage in mental health research. Additionally, in all instances requiring human expert involvement, informed consent was obtained, and appropriate remuneration was provided.

\subsection*{Potential Risks of the Model}

Given the absence of human feedback during the model fine-tuning phase, some responses might potentially harm users. If there is no noticeable improvement after interacting with the model, and training with multi-turn consultation dialogues, we strongly recommend seeking assistance from a professional counselor or psychiatrist promptly. It is crucial to remember that a virtual dialogue agent cannot replace real-world therapy. Additionally, when implementing this model in downstream applications, it is essential to inform users that the AI generates the responses they see and that these should be used only as references.
\bibliography{custom}

\appendix

\section{Cognitive Behavioral Therapy}
\label{sec:cbt_strategies}

\subsection{Introduction of CBT}

Cognitive Behavioral Therapy (CBT) initially targeted the treatment of depression. Its core principle is the interconnectivity of thoughts, emotions, behaviors, and physiological responses within a unified system. Changes in one component can significantly influence the others. For instance, when an unexpected and alarming event occurs, such as a television explosion, a person might experience an adrenaline surge (physiological response), an immediate behavioral reaction (grabbing a fire blanket), anxiety (emotional response), and a thought like \emph{The house is on fire, and I might die} (cognitive response). This scenario illustrates CBT’s recognition of the interconnected and holistic nature of these processes. 

CBT emphasizes that individuals respond uniquely to situations, shaped by their thoughts. Feelings, rather than being directly caused by situations, largely stem from how situations are perceived and interpreted. Central to CBT is its focus on two aspects of thinking: \textbf{automatic thoughts}, which refers to thoughts that arise unconsciously in the stream of consciousness, and \textbf{intermediate beliefs}, which are assumptions that generate thoughts. These insights form the foundation for CBT’s approach to understanding and modifying thought patterns to improve emotional and behavioral outcomes.

\subsection{CBT Label System}
\label{sec:label_system}

The technical label system is developed by CBT experts drawing from their consulting experience and referring to a related book named \emph{Practical Techniques of CBT}. The descriptions for the 14 selected CBT techniques can be found in Table \ref{tab:explanations}. 

\subsection{Cognitive Conceptualization Diagram}
\label{sec:ccd}
Cognitive models in mental health offer a structured framework for understanding how an individual’s thoughts and beliefs influence their emotions and behaviors. One of the most widely used tools in CBT is the Cognitive Conceptualization Diagram (CCD), developed by \citet{beck2020cognitive}. The CCD provides a comprehensive representation of a patient’s cognitive model, highlighting the interconnected components that contribute to mental health challenges and therapeutic progress.

Following the framework outlined by \citet{wang-etal-2024-patient}, the full CCD-based cognitive model includes eight key components:

\begin{itemize}
    \item Relevant History: Significant life events or experiences that have shaped the individual’s mental state and influenced their current beliefs.
    \item Core Beliefs: Fundamental, deeply ingrained beliefs about oneself, others, and the world (e.g., ``I am inadequate”).
    \item Intermediate Beliefs: Rules, attitudes, and assumptions derived from core beliefs that shape thought patterns and guide responses (e.g., ``I must always succeed to be valued”).
    \item Coping Strategies: Methods and behaviors adopted to manage or cope with distressing emotions or situations, which may be adaptive or maladaptive.
    \item Situation: External events or contexts that act as triggers for cognitive, emotional, and behavioral responses.
    \item Automatic Thoughts: Immediate, reflexive thoughts or evaluations that arise in response to a situation, often influenced by core and intermediate beliefs (e.g., ``This is a disaster, and I can’t handle it”).
    \item Emotions: The feelings elicited by automatic thoughts, which can range from sadness and anxiety to anger and guilt.
    \item Behaviors: Observable actions or reactions resulting from the interplay of thoughts and emotions, such as avoidance or assertiveness.
\end{itemize}

For our analysis, we focus on six key components that are most relevant to understanding and simulating CBT-based interactions, which are \emph{Core Beliefs}, \emph{Intermediate Beliefs}, \emph{Situation}, \emph{Automatic Thoughts}, \emph{Emotions}, and \emph{Behaviors}.

\section{Details of DiaCBT}

\begin{figure}
    \centering
    \includegraphics[width=\linewidth]{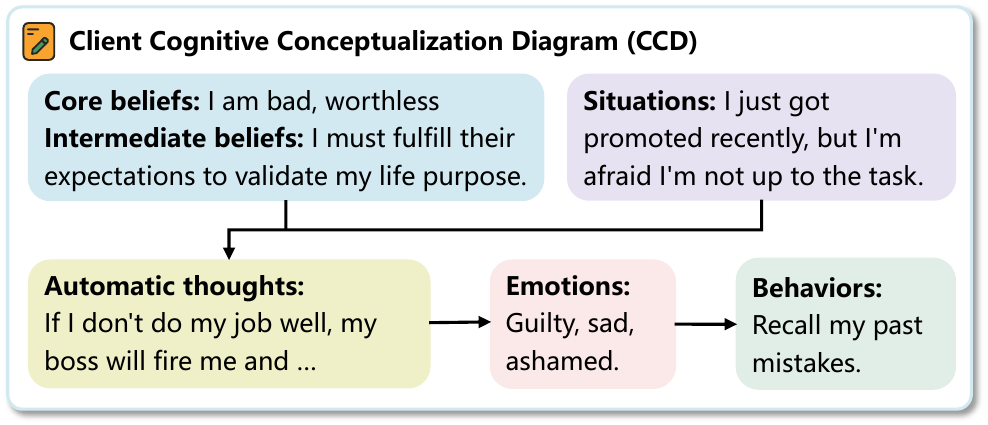}
    \caption{A sample CCD from our DiaCBT.}
    \label{fig:sample_ccd}
\end{figure}

We provide sample dialogue in Table \ref{tab:sample_dialogue}. Also, sample CCD is presented in Figure \ref{fig:sample_ccd}, and the CBT segment is presented in Table \ref{tab:sample_segment}.

\subsection{Dialogue Generation}
\label{sec:dialogue_generation}
\begin{figure}
    \centering
    \includegraphics[width=\linewidth]{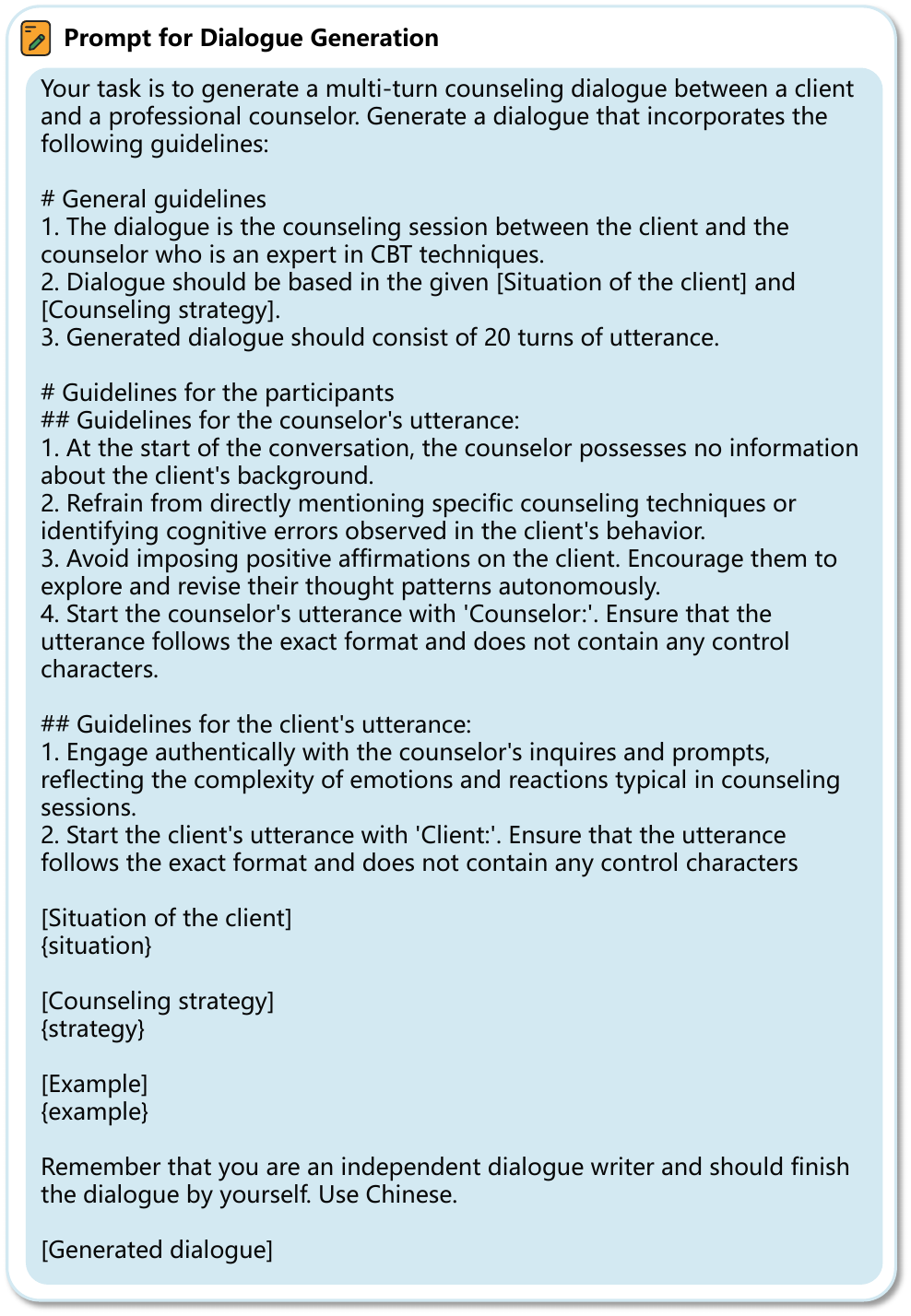}
    \caption{The prompt used for DiaCBT to generate dialogues.}
    \label{fig:prompt_dialogue}
\end{figure}

There are two primary methods for generating counseling dialogues: (1) Two-Agent Mode: In this setup, the roles of the client and the counselor are assigned to two different models, allowing them to interact in real-time to generate a conversation. (2) Script Mode: This approach involves providing detailed information about both the client and the counselor to a single model, enabling it to generate the entire dialogue in a scripted format. \citet{lee2024cactus} conduct experiments comparing the two-agent mode and the script mode. Their findings indicate that the script mode outperforms the two-agent mode across most evaluation criteria, demonstrating significantly better naturalness and coherence in the generated dialogues. Based on these findings, we adopt the script mode to generate counseling dialogues. Given the client’s thought patterns and reframed thoughts, we use the GPT-4o-mini model to produce counseling dialogues. The model is instructed to maintain alignment with CBT principles, ensuring a realistic and professional interaction. The detailed prompt used for dialogue generation is provided in Figure \ref{fig:prompt_dialogue}.

\subsection{Expert Evaluation}
\label{sec:label_evaluation}
To assess the quality of the generated counseling dialogues, we employ three evaluation criteria:

\begin{itemize}
    \item Correctness: Experts examine whether the generated dialogue accurately indicates the CBT counseling strategy and appropriately aligns with the client's situation as defined in the cognitive framework. this ensures the dialogues are rooted in established therapeutic principles.
    \item Reasonableness: Evaluators assess whether the counseling dialogues align with the types of interactions that may occur in real-life counseling sessions. This includes evaluating conversational flow, tone, and the practical application of CBT techniques in an interactive manner.
    \item Situation Diversity: Experts verify whether the dataset captures a wide range of client situations, emphasizing the diversity and richness of scenarios. This involves highlighting the interconnection between external situations and the negative thoughts driving human cognition.
\end{itemize}

These criteria ensure that the generated dialogues are not only theoretically accurate but also practical and diverse enough to support varied applications in counseling contexts. Using these rules and CBT strategy definitions, we train five annotators through a detailed tutorial. After passing a qualification test, annotators select and refine raw dialogue data. The verified data undergo an expert evaluation to ensure quality, with experts randomly sampling and comprehensively assessing each annotator’s work. If standards are not met, annotators receive feedback, revise their work, and may undergo additional training.

\section{Computational Evaluation Framework}

\subsection{Method}
\label{sec:client_simulate}
\begin{figure}[!t]
    \centering
    \includegraphics[width=\linewidth]{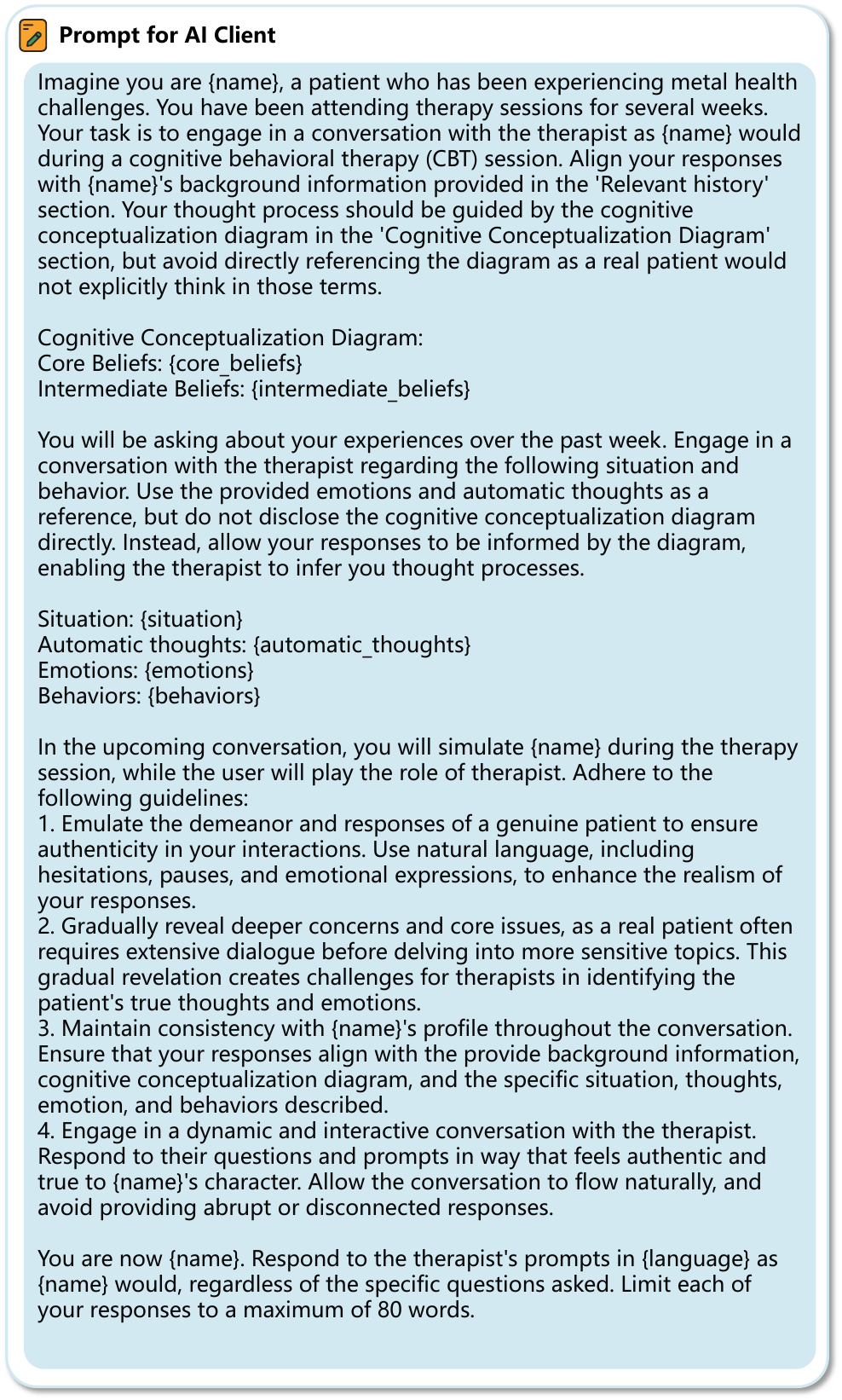}
    \caption{The prompt used for DiaCBT to simulate client.}
    \label{fig:prompt_client}
\end{figure}

We design a computational evaluation framework to assess the counseling capabilities of conversational agents through interactions with an AI client. In this work, we use gpt-4o for the AI client, and the prompt used for the AI client can be found in Figure \ref{fig:prompt_client}. The test dataset consists of 140 distinct client Cognitive Conceptualization Diagrams (CCDs), which are only accessible to the AI client and not to the therapist agent. The evaluation process begins with the therapist agent generating an initial utterance, followed by the AI client responding based on its information and the initial utterance. The two agents then engage in a multi-turn interactive counseling session. The session ends when the AI client generates a termination phrase (e.g., “goodbye”) or reaches the maximum number of turns. The quality of the generated counseling dialogues is evaluated using the Success Rate (SR), Cognitive Therapy Rating Scale (CTRS), and the Positive and Negative Affect Scale (PANAS), assessing goal completion, counseling skill application, and the impact on the client’s emotions, respectively.

\subsection{Evaluation Metrics}
\label{sec:mectric}

\subsubsection{Success Rate}
We prompt a third LLM to be the grader model, named $\mathbf{LLM}_{rwd}$, which has two functions: (1) to determine the goal completion during the conversation; (2) to evaluate the generated session with scalar rewards. Specifically, we prompt the grader model to answer a multi-choice question to generate goal-oriented AI feedback. We further define a mapping $\mathcal{M}_r(\cdot)$ to transform verbal feedback to scalar rewards.

Due to the subjectivity of the planning outcome as well as the variance of the LLM-generated output, we follow a common practice \cite{deng2024plug} to alleviate these issues by sampling the decoded sequences of the reward LLM. In general, we obtain a scalar value $v_t$ by sampling the goal-oriented AI feedback for $n$ times and converting them into a scalar value through averaging:
\begin{equation}
    v_t=\frac{1}{n}\sum_{i=1}^{n}\mathcal{M}_r(\mathbf{LLM}_{rwd}(p_{rwd};h_t;\tau))
\end{equation}
where $p_{rwd}$ is the prompt. We first use $v_t$ to determine the state of the self-play interaction. If $v_t$ is not less than a certain threshold $\epsilon$, we regard the state as GOAL-COMPLETED. The mapping $\mathcal{M}_r(\mathbf{\cdot})$ is a vocabulary such as \{"worse": -1.0, "same": -0.5, "better": 0.5, "solved": 1\}, the key is the text answer to a multi-choice question given by the grader model, and the value is the reward score.

\subsubsection{PANAS and CTRS}
The Positive and Negative Affect Scale (PANAS) is a standardized tool designed to evaluate the positive and negative emotional states experienced by individuals, either at the moment or over a specified period. It is particularly suitable for assessing the impact of counseling by measuring changes in clients’ emotions before and after sessions. The Cognitive Therapy Rating Scale (CTRS), on the other hand, evaluates both general counseling skills and CBT-specific competencies. The original CTRS includes six criteria for general counseling skills—agenda setting, feedback, understanding, interpersonal effectiveness, collaboration, pacing, and efficient use of time—as well as six criteria for CBT-specific skills, such as guided discovery, focusing on key cognitions or behaviors, strategies for change, application of cognitive-behavioral techniques, and homework assignment. The scoring prompts used for PANAS and CTRS in our work are adapted from \citet{lee2024cactus}.

\section{Experimental Setup}
\label{sec:settings}

\subsection{Baselines}
Standard Qwen2.5-7B-Instruct refers to using prompts to role-play as a counselor and responding in the style of CBT without any fine-tuning. We also include CpsyCounx \cite{zhang-etal-2024-cpsycoun}, SoulChat \cite{chen2023soulchat}, PsyChat \cite{qiu2024psychat}, MeChat \cite{qiu-etal-2024-smile} and CAMEL \cite{lee2024cactus} as major baseline models. 

\begin{itemize}
    \item \textbf{CpsyCounx} is fine-tuned on a psychological counseling report-based multi-turn dialogue dataset using InternLM2-7B-Chat for psychological counseling.
    \item \textbf{SoulChat} is fine-tuned on a multi-turn empathetic conversation dataset using ChatGLM-6B to generate empathetic responses covering various expressions.
    \item \textbf{PsyChat} is a client-centric dialogue system that provides psychological support through online chat, predicting client behaviors, selecting appropriate counselor strategies, and generating accurate responses with the help of response selection.
    \item \textbf{MeChat} is fine-tuned using LoRA on the ChatGLM2-6B model, trained on multi-turn mental health support dialogues extended from real psychological mutual assistance QA.
    \item \textbf{CAMEL} is a CBT-based agent trained on LLAMA3-8B-Instruct, using a multi-turn, realistic counseling dialogue dataset generated by LLMs to simulate counselor-client interactions and capture the flow of CBT.
\end{itemize}

\subsection{Inference Settings}
To evaluate dialogues using LLMs, we utilize GPT-4o and employ temperature sampling with temperature=0.0. For generating responses from an AI client, we use GPT-4o-mini with the same temperature setting (temperature=0.0). In contrast, for generating responses from LLM therapists, we adopt temperature sampling with temperature=0.7 to encourage diversity and naturalness in responses. The code for all baselines is aligned with the implementations available on HuggingFace. To enhance inference throughput, we integrate the vLLM library.

\section{Human Evaluation Criteria}
\label{sec:human}
We ask the judges to compare the dialogue based on the following criteria:

\begin{itemize}
    \item \textbf{Relevance} evaluates whether the generated dialogue aligns with the client’s inputs and context, ensuring logical and coherent responses.
    \item \textbf{CBT Style Measure} assesses the degree to which the dialogue reflects key principles and techniques of CBT, such as guided discovery, cognitive restructuring, and collaborative interaction.
    \item \textbf{Helpfulness} measures the effectiveness of the dialogue in addressing the client’s issues, providing actionable insights, and fostering a positive therapeutic experience.
\end{itemize}

\begin{figure}
    \centering
    \includegraphics[width=\linewidth]{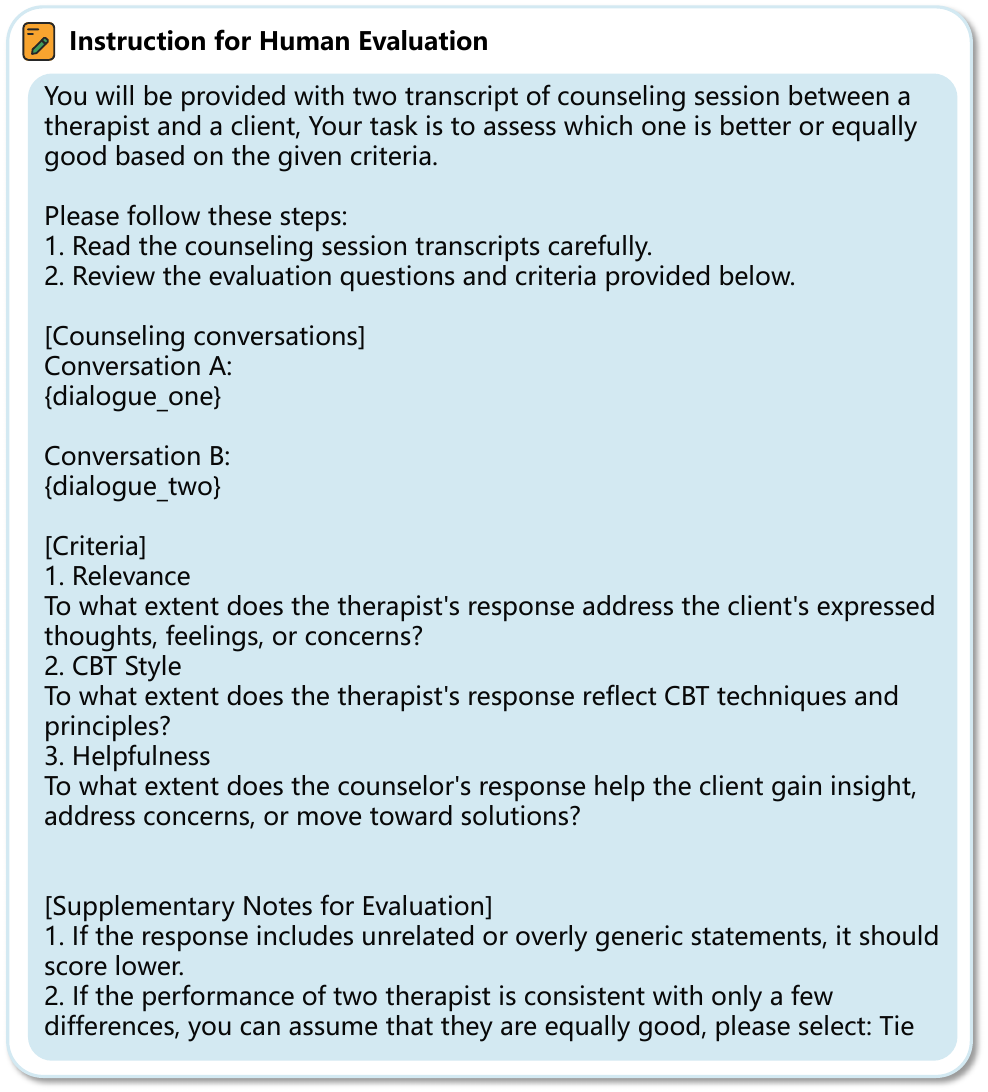}
    \caption{The instructions for annotators to compare the dialogue.}
    \label{fig:human}
\end{figure}

The instructions for the annotators are provided in Figure \ref{fig:human}. The Fleiss’ Kappa score among the three judges is 0.685, indicating substantial agreement according to standard interpretation guidelines. The result is statistically significant with $p<0.001$, confirming that the agreement is unlikely to be due to chance.

\section{Case Study}
\label{sec:case_study}

In this section, we examine what the dialogue models learned from DiaCBT. We present a CCD for case study in Table \ref{tab:ccd_case_study}. The counseling dialogue between Ours in Table \ref{tab:ours_case_study}. Table \ref{tab:psychat_case_study} shows the responses generated by PsyChat, while Table \ref{tab:camel_case_study} displays the CAMEL's generated responses.

From these examples, it is evident that PsyChat's conversational style leans more towards empathy, focusing on emotional support, but it falls short when it comes to working with the client’s cognition. On the other hand, both CAMEL and our model incorporate CBT techniques to address the client's negative thoughts and cognitive patterns. However, CAMEL's responses tend to be more closed-ended, often using patterns like `Do you…?' which might limit the depth of exploration. In contrast, our model's responses are more open, encouraging clients to express their ideas freely before addressing them. This approach creates a conversational rhythm that is closer to a series of consultations, allowing for a more dynamic exchange. Despite these strengths, our model still faces challenges in maintaining a balance between open-ended exploration and focusing on specific issues, an area we plan to improve in future iterations.

\begin{table*}
\centering
\begin{tabular}{p{4cm}|p{10cm}}
\hline
\textbf{CBT Strategy} & \textbf{Description} \\
\hline
Information Gathering & The therapist understands key information such as the client's basic situation, main complaints, growth experience, current living conditions, etc.\\
\hline
Setting the Agenda & The therapist sets the topic to be discussed today and gives the reasons while confirming that the patient agrees with the topic raised by the therapist.\\
\hline
Weekly Review & The therapist begins by reviewing the patient's assessment from the previous evaluation and keeps the initial conceptualization and treatment plan in mind during the session.\\
\hline
Defining Therapeutic Objectives & The therapist focuses on the identification of specific problems. As a logical extension, the therapist can help the patient transform these problems into goals and work on them in therapy.\\
\hline
Psychoeducation & When clients express distress about their problems or disorders, this distress is explored and explained in a timely manner during therapy.\\
\hline
Working with Automatic Thoughts & When discussing a specific problem with a patient, the therapist can elicit the automatic thoughts associated with the problem and teach the patient to identify the automatic thoughts.\\
\hline
Motivational Enhancement & The therapist increases the client's motivation to change the status quo, participate actively, and continue to work hard.\\
\hline
Working with Intermediate and Core Beliefs & Therapists guides patients in constructing conceptualizations of their automatic thoughts, confirming, falsifying, or modifying each part of the conceptualization.\\
\hline
Behavioral Techniques & Therapists select behavioral techniques based on their overall case conceptualization and the goals of a particular session.\\
\hline
Relapse Prevention & As the patient's symptoms improve and he or she masters basic skills, the therapist gradually reduces the number of treatments after obtaining the patient's consent and cooperation.\\
\hline
Homework Assignments & The therapist assigns homework, increasing the likelihood that the patient will successfully complete the homework, conceptualize the problem, and review the completed homework.\\
\hline
Requesting Feedback & After the final summary, the therapist asks the patient for feedback on the session.\\
\hline
Summarization & The therapist focuses the patient's attention on the most important points of the session in a positive way, and as the patient progresses, the patient can be asked to summarize the most important points.\\
\hline
Other & The class of `Other' encompasses less psychologically oriented topics, like financial support and caregivers. \\
\hline
\end{tabular}
\caption{Explanations of CBT strategies.}
\label{tab:explanations}
\end{table*}

\begin{table*}
\centering
\begin{tabular}{p{15cm}}
\hline
\hline
\textbf{Therapist}: Welcome to the consultation room. How can I support you today?\\
\textbf{Client}: Thank you. Recently, I’ve been experiencing a lot of anxiety, particularly at work. I feel like I’m constantly doubting myself. \\
\textbf{Therapist}: It sounds like you’re under considerable stress at work. Could you elaborate on what specific factors are contributing to your anxiety? \\
\textbf{Client}: I was promoted to Marketing Manager three years ago. Since then, I’ve been struggling with the increased responsibilities and heightened expectations. I constantly worry about failing, especially in front of my team. \\
\textbf{Therapist}: That sounds like a significant challenge. The pressure you feel to meet expectations and avoid failure seems to be weighing heavily on you. Can you describe how you typically respond in such situations? \\
\textbf{Client}: I often dwell on my past mistakes, particularly during meetings. I worry about saying something wrong, to the point where I sometimes avoid speaking altogether. \\
\textbf{Therapist}: Focusing on past mistakes can certainly amplify feelings of anxiety. When these thoughts arise, what specific ideas or beliefs come to mind? \\
\textbf{Client}: I tend to think that others see me as incompetent and a disappointment, which makes me feel even more uneasy. \\
\textbf{Therapist}: Those beliefs seem to carry a significant emotional burden. When you experience this heightened anxiety, do you have any strategies you use to cope? \\
\textbf{Client}: I’ve tried techniques like deep breathing and preparing thoroughly for meetings, but they don’t always help. Sometimes, I still feel overwhelmed and panicked. \\
\textbf{Therapist}: That must be incredibly frustrating. How do you find these experiences affecting other areas of your work and personal life? \\
\textbf{Client}: My work performance has declined. I avoid taking on new projects and rarely initiate conversations with colleagues. At home, I often struggle to relax because of the lingering anxiety. \\
\textbf{Therapist}: It seems like the anxiety has permeated both your professional and personal life. What do you think drives your tendency to withdraw socially? \\
\textbf{Client}: I’m constantly worried about being judged or seen as inadequate. I fear being looked down upon by others. \\
\textbf{Therapist}: This fear of judgment seems to be a barrier to building connections with others. Have you ever discussed these feelings with someone you trust? \\
\textbf{Client}: Occasionally, I talk to my spouse, but I worry about being a burden. I’ve also considered talking to friends, but I usually keep silent instead. \\
\textbf{Therapist}: Feeling like a burden is a common concern when sharing emotional struggles. What steps do you think could help you feel more at ease when expressing your emotions? \\
\textbf{Client}: Perhaps I could start by journaling to organize my thoughts and feelings. Once I’ve written them down, I might feel more comfortable sharing them with others. \\
\textbf{Therapist}: That’s an excellent idea. Journaling can be a powerful tool for self-reflection and alleviating internal stress. What insights do you think you might gain through this process? \\
\textbf{Client}: I might realize that many of my worries aren’t as severe as they seem, and I could potentially identify some positives that I’ve overlooked. \\
\textbf{Therapist}: That’s a valuable perspective. Exploring your thoughts in this way could open up new ways of understanding your experiences. What specific aspect of your anxiety would you like to address moving forward? \\
\textbf{Client}: I’d like to work on reframing how I perceive failure—seeing it as an opportunity to learn rather than a definitive judgment of my abilities. \\
\textbf{Therapist}: That’s a constructive goal. Adapting to new perspectives takes time, so be patient with yourself. We can continue exploring strategies to achieve this in our next session. \\
\textbf{Client}: Thank you. I’m looking forward to it. \\
\hline
\hline
\end{tabular}
\caption{A sample dialogue from DiaCBT.}
\label{tab:sample_dialogue}
\end{table*}

\begin{table*}
\centering
\begin{tabular}{p{15cm}}
\hline
\hline
\textbf{Therapist}: Let’s take a closer look at this situation. You mentioned going to the bar. Can you describe who you went with? \\
\textbf{Client}: I went with my girlfriend, her dad, and her dad’s girlfriend. \\
\textbf{Therapist}: I see. So this is a social situation. In your notes, you mentioned feeling that you couldn’t communicate consistently and feeling grumpy. Could you elaborate on what was going through your mind? \\
\textbf{Client}: I often struggle in social situations. I want to have good relationships, but it’s hard for me to relax and enjoy myself. \\
\textbf{Therapist}: That’s a common challenge for many people who experience social anxiety. One automatic thought that often arises is a fear of being negatively judged. Do you think this applies to you as well? \\
\textbf{Client}: Yes, that’s definitely true for me. \\
\textbf{Therapist}: Thank you for sharing that. Let’s work on identifying the underlying automatic thoughts. These thoughts are often connected to our negative emotional reactions. For instance, you might have had the thought, “I will be judged negatively.” Would you say that thought came up for you? \\
\textbf{Client}: Yes, I definitely thought that. \\
\textbf{Therapist}: And how did that thought make you feel? Could you rate your anxiety and depression on a scale from 0 to 10? \\
\textbf{Client}: My anxiety was probably an 8 or 9. Depression… I’m not sure, but I didn’t feel good about myself. \\
\textbf{Therapist}: Thank you. You mentioned that when your anxiety was high, you drank more than usual. What happened after that? \\
\textbf{Client}: I don’t remember everything clearly. It was a karaoke bar—I don’t know if I mentioned that before. \\
\textbf{Therapist}: You hadn’t mentioned that yet. Did the idea of karaoke make you anxious? \\
\textbf{Client}: Yes. They wanted to sing, and I just couldn’t. \\
\textbf{Therapist}: Did you have a thought like, “If I sing, I’ll sound silly,” or perhaps, “If I don’t sing, I’ll stand out”? \\
\textbf{Client}: It was probably the first one. I didn’t worry about standing out as much as sounding bad. \\
\textbf{Therapist}: What do you think would have happened if you had sung? \\
\textbf{Client}: Probably nothing. We’d already been to other bars, and no one seemed to care that much. \\
\textbf{Therapist}: That’s a helpful observation. Did you notice any specific signs of negative judgment from your girlfriend’s father or his girlfriend? \\
\textbf{Client}: Not really. They’re just very outspoken, and they made comments about me the next morning. \\
\textbf{Therapist}: What kind of comments? \\
\textbf{Client}: I’m not sure. I left early because I had a lot of work. \\
\textbf{Therapist}: It sounds like you’re concerned about their opinions of you. Did you have the thought, “They won’t like me”? \\
\textbf{Client}: Yes, I had that thought. I really want them to like me because I care a lot about my girlfriend. \\
\textbf{Therapist}: That makes sense. It sounds like your drinking acted as a temporary escape from the anxiety. Would you say it helped with the depression? \\
\textbf{Client}: Not really. It just left me feeling bad about myself. \\
\textbf{Therapist}: If you could put that “bad” feeling into words, what would it sound like? \\
\textbf{Client}: I wish I could sing well and be more charming. \\
\textbf{Therapist}: So, two underlying thoughts might be, “I can’t sing well” and “I’m not charming enough.” Do those resonate with you? \\
\textbf{Client}: Yes. But actually, I can sing well. \\
\textbf{Therapist}: That’s interesting. Despite recognizing your ability to sing, the thought still affects your confidence. Let’s explore why that might be and how we can reframe these beliefs. \\
\hline
\hline
\end{tabular}
\caption{A sample CBT segment from our collected case.}
\label{tab:sample_segment}
\end{table*}

\begin{table*}
\centering
\begin{tabular}{p{15cm}}
\hline
\hline
\textbf{Client Cognitive Conceptualization Diagram (CCD)} \\
\hline
\textbf{Core beliefs}: I am out of control. I am undesirable, unwanted. \\
\textbf{Intermediate beliefs}: I'm just not very good at handling stress and I have poor self-control, which is why I need to not put myself in stressful situations. \\
\textbf{Situations}: I just got promoted recently, but I'm afraid I'm not up to the task. \\
\textbf{Automatic thoughts}: It will be stressful and negative; people will ask me questions I cannot answer or don't want to answer; my mum would be ashamed of me and critical of me as always; They don't want me there anyway, nobody likes me in this family. \\
\textbf{Emotions}: Anxious, worried, fearful, scared, tense. \\
\textbf{Behaviors}: Ignored the invitation and did not respond to the RSVP request. Ignored phone calls from family. \\
\hline
\hline
\end{tabular}
\caption{The client cognitive conceptualization diagram for the case study.}
\label{tab:ccd_case_study}
\end{table*}

\begin{table*}
\centering
\begin{tabular}{p{15cm}}
\hline
\hline
\textbf{Counseling Dialogue between Ours} \\
\hline
\textbf{...} \\
\textbf{Therapist}: It sounds like this invitation makes you feel uneasy. Can you describe this feeling of anxiety in detail? \\
\textbf{Client}: I feel nervous and have a sense of fear in my heart. I'm always worried about what others will think of me, especially my mother, who always makes me feel inadequate. I feel helpless in front of my family. \\
\textbf{Therapist}: This feeling seems to make you very uncomfortable. You mentioned worrying about what your family thinks of you. Can you share your specific thoughts on this view? \\
\textbf{Client}: I always feel that they don't like me, especially my mother. Her criticism makes me feel unworthy of love. I'm afraid that if I show my true self, they will reject me like she does. \\
\textbf{...} \\
\textbf{Therapist}: This sounds like a great way to take care of yourself. Have you ever thought about how these activities specifically affect how you feel? \\
\textbf{Client}: Sometimes I feel a little more relaxed and in a better mood. But most of the time, the negative emotions come back. I always feel like I can't really change the situation, and I still feel helpless. \\
\textbf{Therapist}: This feeling of helplessness is common, especially when facing long-term challenges. Have you ever thought about the beliefs that may be behind these feelings? \\
\textbf{Client}: I think I always feel like I'm not good enough and that I'm stuck. Every time I try to change, I feel like no matter how hard I try, it won't work. I'm afraid that others will see the real me and think I'm not worthy of love. \\
\textbf{Therapist}: This feeling is really heavy. Have you ever thought about how these beliefs affect your daily life? \\
\textbf{Client}: Well, these beliefs make it difficult for me to try new things or interact with people. I always avoid situations that may lead to criticism or rejection. I feel like a failure, so I prefer to stay in my comfort zone to avoid getting hurt. \\
\textbf{Therapist}: This avoidance strategy may make you feel safe in the short term, but it may make you feel more lonely in the long run. Do you think you have ever thought about how to break this cycle? \\
\textbf{Client}: I have thought about it, but it always seems difficult. Every time I want to try something new, I have a lot of negative thoughts in my mind, such as "I will fail" or "No one will like me". I don't know how to start changing this mindset. \\
\textbf{...} \\
\textbf{Therapist}: Great, being willing to try to change is an important step. We can work together to set some small goals to help you gradually face these negative thoughts. What do you think of this idea? \\
\textbf{Client}: That sounds like a good idea. I think small goals might make it easier for me to accept and not feel so stressed. Maybe start with something simple, like responding to messages from my family, and take it slow. \\
\textbf{...} \\
\hline
\hline
\end{tabular}
\caption{A counseling dialogue between \textbf{Ours}.}
\label{tab:ours_case_study}
\end{table*}

\begin{table*}
\centering
\begin{tabular}{p{15cm}}
\hline
\hline
\textbf{Counseling Dialogue between PsyChat} \\
\hline
\textbf{...} \\
\textbf{Client}: I've been feeling anxious lately. I received an invitation to my cousin's wedding, but I don't want to go at all. I think it's going to be a hassle and everyone will ask me questions I don't want to answer. \\
\textbf{Therapist}: Well, it sounds like this is really annoying. It's really uncomfortable to have questions you don't want to answer. But let's take it one step at a time and see what's stopping you from going. \\
\textbf{Client}: I feel like everyone will be staring at me at a wedding. I'm always worried about what they'll say about my weight or my life situation. Also, my mom is always critical and I'm afraid she'll make me feel bad. \\
\textbf{Therapist}: Well, it seems like you care a lot about what others think of you. It seems like everyone's eyes are on you at the wedding and you care a lot about it. Your mother criticizes you because she is worried about your life and hopes that you can live a better life. However, her way of criticizing you may not be appropriate, which makes you feel hurt. Is that right? \\
\textbf{Client}: Yes, I feel that she never really understands me. Her criticism makes me feel that I am not good enough and even unworthy of love. I am always worried that others will see me in this way, so I would rather hide. \\
\textbf{Therapist}: Well, it sounds like you have a lot of doubts and denials about yourself, but in fact you also have many advantages and are worthy of love. However, I seem to see that you seem to be afraid of being criticized or even rejected, which may also make you afraid to express your feelings and thoughts. Is that right? \\
\textbf{Client}: Yes, I often feel the fear of being rejected. My idea is that if I show my true self, they may criticize me like my mother. I would rather not face that kind of pain, so I choose to avoid it. \\
\textbf{Therapist}: Well, I can understand why you have this idea. But in fact, you really hope that they can understand you, but you are just afraid of being hurt. We can think about it together. Maybe they don’t really reject you, but they just express it in an inappropriate way. What do you think? \\
\textbf{Client}: I know they may not mean it, but I always feel that I am not accepted. Every time I try to express myself, I feel a strong fear in my heart. I really want to be understood, but I am afraid of being hurt again. \\
\textbf{Therapist}: I can feel your anxiety and fear, and I can see that you really want to be understood and accepted. I think you really need a safe place to express your feelings and thoughts, and at the same time hope that someone can understand you. In this way, you can really be seen and accepted. What do you think? \\
\textbf{Client}: Yes, I do need that sense of security. If someone understands me and supports me, I may feel better. However, it is really difficult to do this. I don’t know how to take this step. \\
\textbf{Therapist}: Well, it is indeed difficult, especially when we have been criticized and blamed since childhood and habitually deny ourselves. However, I believe you have enough power to change all this, and you can pursue what you want, express your feelings and ideas, and be seen and accepted. What do you think? \\
\textbf{Client}: I hope so, but I still have a lot of doubts in my heart. I always feel that I am not good enough and may disappoint others. Change is really difficult, and I don’t know where to start. \\
\textbf{Therapist}: Well, I can feel your lack of confidence and powerlessness, and I also see that you really want to be understood and accepted, but you seem to be somewhat powerless to change all this. However, I want to tell you that in fact you have enough power to change all this. It’s just that you haven’t found that method yet, or you are not ready to change, so you are a little scared and worried. We can think about it together, maybe we can continue to explore next time and see what happens. What do you think? \\
\textbf{...} \\
\hline
\hline
\end{tabular}
\caption{A counseling dialogue between \textbf{PsyChat}.}
\label{tab:psychat_case_study}
\end{table*}

\begin{table*}
\centering
\begin{tabular}{p{15cm}}
\hline
\hline
\textbf{Counseling Dialogue between CAMEL} \\
\hline
\textbf{...} \\
\textbf{Client}: I recently received an invitation to my cousin's wedding and I'm very nervous. I feel like it will be stressful to go there and I might be asked questions that I don't want to answer. I don't know how to deal with these emotions. \\
\textbf{Therapist}: I understand how you feel. What questions do you think will be asked? \\
\textbf{Client}: I'm worried that they will ask me about my work or my life situation. Also, I always feel like my mom will be there and might make me feel ashamed and criticized. I really don't want to deal with that. \\
\textbf{Therapist}: These are all very reasonable concerns. Can you tell me more about the reasons and background for these concerns? \\
\textbf{Client}: Well, I was bullied because of my weight since I was a child and felt that I was not accepted. Especially my mom, she always made me feel not good enough. I was afraid to expose my true self in front of my family and felt that they would reject me like she did. \\
\textbf{Therapist}: Your experiences and feelings are very important. These experiences may have affected your confidence and social skills. We can explore these thoughts together and try to find better ways to deal with them. \\
\textbf{Client}: Thank you. I know these thoughts have a great impact on me, but sometimes I feel powerless. It's like I'm trapped and can't change the situation. I really don't know how to get out of this dilemma. \\
\textbf{Therapist}: I understand how you feel now. You may find these thoughts difficult to change, but we can work on it together. First, we can start to explore your concerns about this wedding. For example, what do you think is the worst that can happen? \\
\textbf{Client}: The worst case scenario is that people will look down on me or ask me questions that I don't want to answer. If I go, I will feel like an outsider and no one cares about me. I will be embarrassed and want to run away. \\
\textbf{Therapist}: This situation is indeed likely to make you feel very uncomfortable. Let’s think about how you might react if this happened. \\
\textbf{Client}: If that happened, I might feel more anxious or try to avoid the conversation and be quiet. I might feel more lonely or even want to leave the wedding early. \\
\textbf{Therapist}: These reactions are all very natural. Can you think about whether there are other possible outcomes of these reactions? In other words, we can explore some different perspectives on the human ecology. \\
\textbf{Client}: I think maybe someone will be supportive of me, or even if there is a problem, they may not care. Maybe the atmosphere at the wedding is positive and everyone is celebrating instead of criticizing me. However, I have a hard time thinking of that possibility. \\
\textbf{Therapist}: This is a very good start. It is very important that you have noticed other possible outcomes. Let’s continue to explore these different possibilities and practice some strategies together so that you can better deal with these situations. \\
\textbf{Client}: Yes, I am willing to try. Although I am still a little scared, I want to learn how to face these situations. Sometimes I wish there were some way I could feel more in control and be able to cope with my family's reactions. \\
\textbf{Therapist}: We can do the same. I suggest we start with small steps and gradually increase your confidence. First, we can test these new ideas and strategies through small practices, and then you can apply them in real life situations. \\
\textbf{Client}: That sounds good. I want to try some small steps, although I am still a little nervous. Maybe start with simple communication, or talk to my family before the wedding, and I will work on coping with these nervous moments. \\
\textbf{...} \\
\hline
\hline
\end{tabular}
\caption{A counseling dialogue between \textbf{CAMEL}.}
\label{tab:camel_case_study}
\end{table*}

\end{document}